\documentclass{article}

\usepackage{microtype}
\usepackage{graphicx}
\usepackage{subfigure}
\usepackage{booktabs} 

\usepackage{hyperref}

\usepackage[accepted]{icml2023}

\usepackage{amsmath}
\usepackage{amssymb}
\usepackage{mathtools}
\usepackage{amsthm}

\usepackage[capitalize,noabbrev]{cleveref}

\theoremstyle{plain}
\newtheorem{theorem}{Theorem}[section]

\newtheorem{lemma}[theorem]{Lemma}
\newtheorem{corollary}[theorem]{Corollary}
\theoremstyle{definition}
\newtheorem{definition}[theorem]{Definition}
\newtheorem{assumption}[theorem]{Assumption}
\theoremstyle{remark}

\usepackage[textsize=tiny]{todonotes}

\icmltitlerunning{Submission and Formatting Instructions for FL-ICML 2023}

\begin{document}

\twocolumn[
\icmltitle{Private Federated Learning with Dynamic Power Control via \\ Non-Coherent Over-the-Air Computation}



\icmlsetsymbol{equal}{*}

\begin{icmlauthorlist}
\icmlauthor{Anbang Zhang}{sch,lab}
\icmlauthor{Shuaishuai Guo}{sch,lab}
\icmlauthor{Shuai Liu}{sch}
\end{icmlauthorlist}

\icmlaffiliation{sch}{School of Control Science and Technology, Shandong University, Jinan, China.}
\icmlaffiliation{lab}{Shandong Key Laboratory of Wireless Communication Technologies, Jinan, China.}

\icmlcorrespondingauthor{Shuaishuai Guo}{shuaishuai\_guo@sdu.edu.cn}

\icmlkeywords{Federated Learning, ICML}

\vskip 0.3in
]



\printAffiliationsAndNotice{}  

\begin{abstract}
To further preserve model weight privacy and improve model performance in Federated Learning (FL), FL via Over-the-Air Computation (AirComp) scheme based on dynamic power control is proposed. The edge devices (EDs) transmit the signs of local stochastic gradients by activating two adjacent orthogonal frequency division multiplexing (OFDM) subcarriers, and majority votes (MVs) at the edge server (ES) are obtained by exploiting the energy accumulation on the subcarriers. Then, we propose a dynamic power control algorithm to further offset the biased aggregation of the MV aggregation values. We show that the whole scheme can mitigate the impact of the time synchronization error, channel fading and noise. The theoretical convergence proof of the scheme is re-derived.
\end{abstract}

\section{Introduction}
\label{submission}

With the substantial increase in computation ability and storage capacity of modern intelligent terminals, distributed FL is utilized the most widely, which provides a promising learning paradigm for the current privacy computation. By pushing model training locally \cite{mcmahan2023communicationefficient}, FL is able to build global models without directly sharing data. This mechanism largely protects the privacy and security \cite{9562559} of users and addresses the potential for data leakage in the context of large amounts of data.

However, there are still numerous key challenges in deploying practical FL applications in the real world due to resource constraints \cite{9210812} and privacy concerns in wireless networks. Several works \cite{8835269} show that if the model parameters or gradients exchanged between EDs and  ES are attacked, sensitive information about local data is still exposed. In addition, a large number of model parameters need to be repeatedly transmitted over wireless channels \cite{9084352}, thus requiring huge communication resources, which is a significant bottleneck.

\subsection{Related Work}
\textbf{Gradient compression:} To overcome the above problems, a prospective solution is gradient quantization, such as SignSGD, QSGD \cite{10.5555/3294771.3294934}, and cpSGD \cite{agarwal2018cpsgd}. But transmitting the gradients can also have some privacy leakage \cite{zhu2019deep}, so deeper compression methods should be considered more. In \cite{bernstein2018signsgd}, it is shown that actually gradients are really useful in terms of the direction rather than size. Thus, SignSGD considers to quantize gradients, achieving a 32 times data compression. Also, based on the FL scheme, it is difficult to recover the data information used in the model by hijacking the gradient direction due to reducing the transmission of information \cite{akoun2022signsgd}, thus ensuring the privacy and security of the data information.

\begin{table}[t]
\caption{The communication cost of different gradient compression
schemes, when training a D-dimensional model with M EDs.}
\label{sample-table}
\vskip 0.15in
\begin{center}
\begin{small}
\begin{sc}
\begin{tabular}{lcccr}
\toprule
ALGORITHM & BITS PER ITERATION \\
\midrule
SGD    & 64MD & \\
QSGD & (2 + log(2M+ 1))MD & \\
TERNGRAD & (2 + log(2M+ 1))MD & \\
SIGNSGD WITH MV   & 2MD &\\
\bottomrule
\end{tabular}
\end{sc}
\end{small}
\end{center}
\vskip -0.1in
\end{table}
 
\textbf{Over-the-air Computation:} The second option to consider is to adopt the weight superposition over the air, i.e., AirComp \cite{9095231}. This combination of communication and computation reduces the latency and bandwidth requirements \cite{6557530}. However,  the usual AirComp scheme requires channel state information (CSI) at the EDs or the ES. To alleviate the channel estimation burden, \cite{9272666} considered that EDs use truncated-channel inversion (TCI) to transmit orthogonal phase shift keying (QPSK) symbols on the orthogonal frequency division multiplexing (OFDM) subcarrier instead of transmitting the gradient or the direction of the gradient directly, which corresponds to a double layer of privacy encryption protection and guarantees absolute security of model weight privacy during communication. 

\textbf{Dynamic Power Control:} In the context of the above approach, the privacy security as well as communication overhead issues are effectively addressed, but the multi-user parameter aggregation produces discrepancies on training process of the global model, which has become the focus of the research. Therefore, we consider to adjust transmitting power \cite{li2022transmit} with EDs in order to offset the impact of the user parameters on the whole model in the opposite direction of the model convergence during the parameter aggregation, thus achieving better model accuracy.

\subsection{Contributions}
All the contributions can be listed as follows:

\textbf{(1)} By considering FL training based on SignSGD with Majority Vote (MV), an Over-the-Air Computation scheme is utilized in which symbols (i.e., directions) of random gradients are transmitted by using OFDM symbols. The MV is obtained by energy detection on the ES, so that CSI is not required at the EDs and ES. 

\textbf{(2)} We design a dynamic power control scheme that trades off MV and the transmitting signs of each ED to offset the effect of the EDs parameters on the whole model in the opposite direction of the model convergence when the parameters are aggregated. 

\textbf{(3)} Then, we re-derive the theoretical convergence proof of the proposed scheme, when MVs are obtained by employing the FL scheme. The experiment results can prove that the proposed scheme is robust to time synchronization errors because it does not encode signs of local stochastic gradients into the phase of the transmitted symbols. 

A figure demonstrating the training process is given in Appendix A.

\section{Overview of Our Approach}
\subsection{Federated Learning With Majority Vote}
We consider an FL system comprising a single edge server coordinating the learning process across $\emph{M}$ EDs. The aim of FL can be represented as finding an optimal model parameter vector $\mathbf{w}^{*}$ that minimizes $F(\mathbf{w})$, i.e.,
\begin{equation}
\mathbf{w}^{*} = \min _{\mathbf{w} \in \mathbb{R}^{q}} F(\mathbf{w})=\min _{\mathbf{w} \in \mathbb{R}^{q}} \frac{1}{|D|} \sum_{\forall(x, y) \in D} f(\mathbf{w}; \mathbf{x}, y),
\end{equation}
where $\mathcal{D}=\bigcup_{m=1}^{M}\left\{\mathcal{D}_{m}\right\}$ is the global dataset set and $f(\mathbf{w}, \mathbf{x}, y)$ is the sample-wise loss function indicating the prediction error, for example, $(\mathbf{x}, y)$ with the FL model parameters $\mathbf{w}=[w_{1}, \ldots, w_{q}]^{\mathrm{T}} \in \mathbb{R}^{q}$ , and $\emph{q}$ is the number of model parameters. 

In the training process, EDs exploit a mini-batch stochastic gradient descent method to calculate local gradient $\tilde{\mathbf{g}}_{m}^{(n)} \triangleq[\tilde{g}_{m, 1}^{(n)}, \ldots, \tilde{g}_{m, q}^{(n)}]^{\mathrm{T}}$ with respect to the current received global model $\mathbf{w}^{(n)}$ as
\begin{equation}
\tilde{\mathbf{g}}_{m}^{(n)}=\nabla F_{m}\left(\mathbf{w}^{(n)}\right)=\frac{1}{d_{\mathrm{b}}} \sum_{\forall\left(\mathbf{x}_{\ell}, y_{\ell}\right) \in \tilde{\mathcal{D}}_{m}} \nabla f\left(\mathbf{w}^{(n)}, \mathbf{x}_{\ell}, y_{\ell}\right),
\end{equation}
where $\tilde{\mathcal{D}}_{m} \subset\mathcal{D}_{m}$ is selected data batch from local data set and $d_{\mathrm{b}}=|\tilde{\mathcal{D}}_{m}|$ as the batch size. In the context of FL processing, SignSGD with the majority vote approach \cite{bernstein2018signsgd} is investigated to solve above problems. The trained real stochastic gradients are converted into sign values by one-bit quantization scheme, which are denoted as $\bar{\mathbf{g}}_{m,i}^{(n)}\triangleq\operatorname{sign}(\tilde{\mathbf{g}}_{m, i}^{(n)})$. Thus, the parameter MV of the $\emph{i}$th global gradient estimate at the ES would be enforced as follows: 
\begin{equation}
v_{i}^{(n)} = \operatorname{sign}\left(\sum_{m=1}^{M} \bar{\mathbf{g}}_{m, i}^{(n)}\right).
\end{equation}
Afterwards, the ES pushes $\mathbf{v}^{(n)}=[v_{1}^{(n)}, \ldots, v_{q}^{(n)}]^{\mathrm{T}}$ to the EDs, and the models at the EDs are updated as:
\begin{equation}
\mathbf{w}^{(n+1)}=\mathbf{w}^{(n)}-\eta \mathbf{v}^{(n)}.
\end{equation}
This procedure is repeated consecutively until a predetermined
convergence criterion is achieved. Combining above process, the corresponding scheme is Algorithm 1.
\begin{algorithm}[tb]
   \caption{signSGD-MV based on AirComp}
   \label{alg:example}
\begin{algorithmic}
   \STATE {\bfseries Input:} learning rate $\eta$, current received global model $\mathbf{w}^{(n)}$, $\emph{M}$ EDs each with $\bar{\mathbf{g}}_{m}^{(n)}$, initialize $\mathbf{w}^{(0)}$.
   \REPEAT 
   \STATE 
   \textbf{On Each Edge Device:}
   
   calculate the sign $\bar{\mathbf{g}}_{m}^{(n)}$ of stochastic gradients.
   
   update $\mathbf{w}^{(n+1)}=\mathbf{w}^{(n)}-\eta \mathbf{v}^{(n)}$.
   
   \textbf{On Edge Server:}

   pull the sign $\bar{\mathbf{g}}_{m}^{(n)}$ from $\emph{m}$ ED with  non-coherent energy detection via AirComp.
   
   broadcast $\mathbf{v}^{(n)}=\operatorname{sign}(\sum_{m=1}^{M} \bar{\mathbf{g}}_{m}^{(n)})$ to all the EDs.
   \UNTIL{reach convergence}
\end{algorithmic}
\end{algorithm}

\subsection{Transmitter Design - Dynamic Power Control on FSK-MV}
We concentrate on uplink communication process based on the MV with AirComp (i.e., FSK-MV \cite{sahin2021distributed}). At the $\emph{n}$th communication round, the superposed symbol on the $\emph{l}$th subcarrier of $\emph{m}$th OFDM symbol can be given by:
\begin{equation}
{\mathbf{y}}_{l, s}^{(n)}=\sum_{m=1}^{M} \sqrt{P_{m}^{(n)}} {\mathbf{H}}_{m, l, s}^{(n)} t_{m, l, s}^{(n)}+{\mathbf{n}}_{l, s}^{(n)},
\end{equation}
where ${\mathbf{H}}_{m, l, s}^{(n)}\in{\mathbb{C}}$ is the channel coefficient with identical Rayleigh distribution, and $\emph{t}_{m, l, s}^{(n)}\in{\mathbb{C}}$ is the transmitted symbol, and ${\mathbf{n}}_{l, s}^{(n)}$ is an additive white Gaussian noise vector on the $\emph{l}$th subcarrier for $\emph{l}\in\{0,1, \ldots, A-1\}$ and $\emph{s}\in\{0,1, \ldots, S-1\}$. The EDs perform a low-complexity operation that each ED activates one of the two adjacent subcarriers determined by the time-frequency index pairs to transmit the signs of the gradients. To express this encoding operation rigorously, let $\emph{f}$ be a bijective function that maps $i \in\{1,2, \ldots, q\}$ to the distinct pairs $\left(s^{+}, l^{+}\right)$ and $\left(s^{-}, l^{-}\right)$. Thus, the $\emph{m}$th ED determines the following bins of modulation symbol ${\mathbf{t}}_{m,l^{+},s^{+}}^{(n)}$and ${\mathbf{t}}_{m,l^{-},s^{-}}^{(n)}$ , $\forall i$, as
\begin{equation}
t_{m, l^{+}, s^{+}}^{(n)}=\sqrt{E_{0}} s_{m, i}^{(n)}\mathbb{I}\left[\bar{\mathbf{g}}_{m,i}^{(n)}\triangleq\operatorname{sign}(\tilde{\mathbf{g}}_{m, i}^{(n)})=1\right],
\end{equation}
and
\begin{equation}
t_{m, l^{-}, s^{-}}^{(n)}=\sqrt{E_{0}} s_{m, i}^{(n)}\mathbb{I}\left[\bar{\mathbf{g}}_{m,i}^{(n)}\triangleq\operatorname{sign}(\tilde{\mathbf{g}}_{m, i}^{(n)})=-1\right],
\end{equation}
where $\emph{E}_{0}=2$ is a factor to normalize the OFDM symbol energy, $\emph{s}_{m, i}^{(n)}$ is a randomization symbol on the unit circle, and $\mathbb{I}$ is the indicate function. As a special case of the mapping function $\emph{f}$, if $\emph{s}^{-}=\emph{s}^{+}$ and $\emph{l}^{-}=\emph{l}^{+}+1$, which holds for all $\emph{i}$, then the adjacent subcarriers of $\emph{m}^{+}$th OFDM symbol is used for the voting scenario, which corresponds to frequency-shift keying (FSK) on the OFDM subcarriers.

\textbf{Dynamic Power Control:} Based on the above analysis of the communication process, we consider the provision of a larger proportion of biased gradient symbols and larger EDs with Gaussian white noise. Therefore, we formulate the dynamic power control design problem in the direction of convergence of the balanced global model in the following form as:
\begin{equation}
{P}_{m}^{(n)}={P}_{m}^{(n-1)} + \left|\frac{1}{q} \sum_{i=1}^{q}\left[\mathbb{I}_{\bar{g}_{m, i}^{(n)}=v_{i}^{(n)}}-\mathbb{I}_{\bar{g}_{m, i}^{(n)} \neq v_{i}^{(n)}}\right]\right|,
\end{equation}
where $v_{i}^{(n)}$ and $\bar{g}_{m, i}^{(n)}$ is form of model parameter aggregation when channel transmission is not considered and the $\emph{i}$th parameter of the vector shared by $\emph{m}$th ED, respectively. Specially, the initialized transmission power is ${P}_{m}^{(0)} = 1$. The above power control scheme is adopted to be able to achieve the convergence direction convergence with the global model. See the proof in Appendix F.

\subsection{Receiver Design - Non-coherent Energy Detection}

Based on above theory, we choose to use a non-coherent energy detection \cite{9839037} approach to obtain the MV. At the receiver ES, we first identify the pairs $\left(s^{+}, l^{+}\right)$ and $\left(s^{-}, l^{-}\right)$, and observe the superposed symbols, which are expressed as:
\begin{equation}
r_{l^{+}, s^{+}}^{(n)}=\sqrt{E_{0}} \sum_{\forall m, \bar{g}_{m, i}^{(n)}=1} \sqrt{{P}_{m}^{(n)}} h_{m, l^{+}, s^{+}}^{(n)} s_{m, i}^{(n)}+n_{l^{+}, s^{+}}^{(n)},
\end{equation}
and
\begin{equation}
r_{l^{-}, s^{-}}^{(n)}=\sqrt{E_{0}} \sum_{\forall m, \bar{g}_{m, i}^{(n)}=-1} \sqrt{{P}_{m}^{(n)}} h_{m, l^{-}, s^{-}}^{(n)} s_{m, i}^{(n)}+n_{l^{-}, s^{-}}^{(n)}.
\end{equation}
Subsequently, we exploit an energy detector to obtain the MV for the $\emph{i}$th gradient as
\begin{equation}
v_{i}^{(n)}=\operatorname{sign}\left(\Delta_{i}^{(n)}\right),
\end{equation}
where $\Delta_{i}^{(n)}$ is the sum energy to detect the votes and $\Delta_{i}^{(n)} = \mathbf{e}_{i}^{+} - \mathbf{e}_{i}^{-}$, whose derivation process is shown in Appendix B. Also, $e_{i}^{+}$ and $e_{i}^{-}$ are the energies of the superposed symbols on adjacent subcarriers. 

\section{Error Probability Analysis and Convergence Rate Performance}
To facilitate subsequent analysis of convergence as well as error probability, several standard assumptions \cite{9844173} are shown as Appendix C.

\subsection{Error Probability Analysis}
\textbf{Received Signal Power of MV:} The above scheme determines the correct MV by comparing $e_{i}^{+}$ and $e_{i}^{-}$ directly. Also, let $\emph{M}_{i}^{+}$ and $\emph{M}_{i}^{-}$ be the number of EDs that vote for $\bar{g}_{m, i}^{(n)}=1$ and $\bar{g}_{m, i}^{(n)}=-1$, respectively. Then, we obtain the expressions of the average received signal power as $\mu_{i}^{+}$ and $\mu_{i}^{-}$ with the following lemma:
\begin{lemma}
For the given $\emph{M}_{i}^{+}$ and $\emph{M}_{i}^{-}$, $\mu_{i}^{+}$ and $\mu_{i}^{-}$ can be calculated as
\begin{equation}
\mu_{i}^{+} \triangleq \mathbb{E}\left[e_{i}^{+}\right]=E_{0} M_{i}^{+} \vartheta+\sigma_{\mathrm{n}}^{2},
\end{equation}
and
\begin{equation}
\mu_{i}^{-} \triangleq \mathbb{E}\left[e_{i}^{-}\right]=E_{0} M_{i}^{-} \vartheta+\sigma_{\mathrm{n}}^{2},
\end{equation}
\end{lemma}
respectively. The parameter $\vartheta$ is the average value of transmission power for all $\emph{i}$, which is equivalent to a constant. The proof is given in Appendix B.

\textbf{Bit Error Probability Analysis:} Based on several assumptions provided in Appendix C, we proceed to analyze the error probability in the sign aggregation process. According to the Appendix F, the performance of our scheme is bounded by probability of misidentifying the correct sign for the $\emph{i}$th gradient, which is obtained by:
\begin{equation}
P_{i}^{\mathrm{err}} \triangleq \operatorname{Pr}\left(\operatorname{sign}\left(\Delta_{i}^{(n)}\right) \neq \operatorname{sign}\left(g_{i}^{(n)}\right)\right),
\end{equation}
which is determined by the level of noise introduced by the data-stochasticity and wireless channel. For a more accurate formalization, $P_{i}^{\mathrm{err}}$ for MV can be obtained as follows:
\begin{lemma}
(Error probability for MV).
Under the dynamic power control method, we derive that stochasticity-induced error in the wireless fading channel for all $\emph{i}$ is bounded as:
\begin{equation}
P_{i}^{err} \leq \frac{\frac{K}{2} \cdot \sqrt{2} / (3 R_{i})}{K+2 / \beta}+\frac{1 / \beta}{K+2 / \beta},
\end{equation}
\end{lemma}
where $R_{i}=\sqrt{d_{b}} \frac{|g_{i}^{(n)}|}{\sigma_{i}}$ is defined as the gradient-signal-to-data-noise ratio \cite{9272666}. Also, the resultant gradient variance reduces from ${\sigma_{i}^{2}}$ to ${\sigma_{i}^{2}}/{d_{b}}$ according to Assumption C.5 and Equation (2). We provide the proof in Appendix D. And Lemma 3.2 implies the following results:
\begin{corollary}
(Legitimate EDs). For $q_{i} < p_{i}$, $X$ must be larger than $K/2$, meanwhile must satisfying $P_{i}^{err} < 1/2$.
\end{corollary}

\subsection{Convergence Rate over Fading Channel}

It is obvious that the proposed scheme maintains the convergence of the original MV. Based on this, we re-derive the theoretical convergence performance of
the proposed scheme as follows:
\begin{theorem}
Consider a FL system based on the proposed scheme, for the mini-batch size $d_{\mathrm{b}}=N / \gamma$ and the learning rate $\eta=1 / \sqrt{\|\mathbf{L}\|_{1} d_{\mathrm{b}}}$, the convergence rate in fading channel is given by
\begin{equation} 
\begin{split}
\mathbb{E}\left[\frac{1}{N} \sum_{n=0}^{N-1}\left\|\mathbf{g}^{(n)}\right\|_{1}\right] \leq \frac{1}{\sqrt{N}}&\Big(\tau \sqrt{\|\mathbf{L}\|_{1}}\left(F(\mathbf{w}^{(0)})-F^{*}+\frac{\gamma}{2}\right)\\&+\frac{2 \sqrt{2}}{6} \sqrt{\gamma}\|\boldsymbol{\sigma}\|_{1}\Big),
\end{split}
\end{equation}
\end{theorem}
where $\gamma$ is a positive integer, $\tau\!=\!\left(1+\frac{2}{\beta K}\right) \frac{1}{\sqrt{\gamma}}$ , and $\beta \triangleq \frac{E_{\mathrm{0}}\vartheta}{\sigma_{\mathrm{n}}^{2}}$. More details of the convergence analysis are shown in Appendix F. The detailed convergence rate analysis can be found in Appendix G.

\section*{Acknowledgements}

The work is supported in part by the National Natural Science Foundation of China under Grant 62171262; in part by Shandong Provincial Natural Science Foundation under Grant ZR2021YQ47; in part by the Taishan Young Scholar under Grant tsqn201909043; in part by Major Scientific and Technological Innovation Project of Shandong Province under Grant 2020CXGC010109.

\section{Experiments}

For experiments, we investigate benchmark image dataset: MNIST. We run our experiments with 31 normal EDs, and partition the training dataset according to the labels. For fair comparison, we set the same hyper-parameters (batch size as 128, local epoch as 1, and learning rate as 0.004). We compare our proposed scheme with two baseline algorithms: SignSGD, FedAvg and FSK-MV to obtain performance results.

In FSK-MV and the proposed algorithm, we configure the arrival time of the EDs signal to be different. In the experiments on the MNIST dataset, we compare in independent identically distributed (iid) as well as non-independent identically distributed (non-iid) data to demonstrate more precisely the performance advantages and disadvantages of the algorithm compared to existing designs.

\begin{figure}[ht]
\vspace{-2.0em}
\vskip 0.2in
\begin{center}
\centerline{\includegraphics[width=\columnwidth]{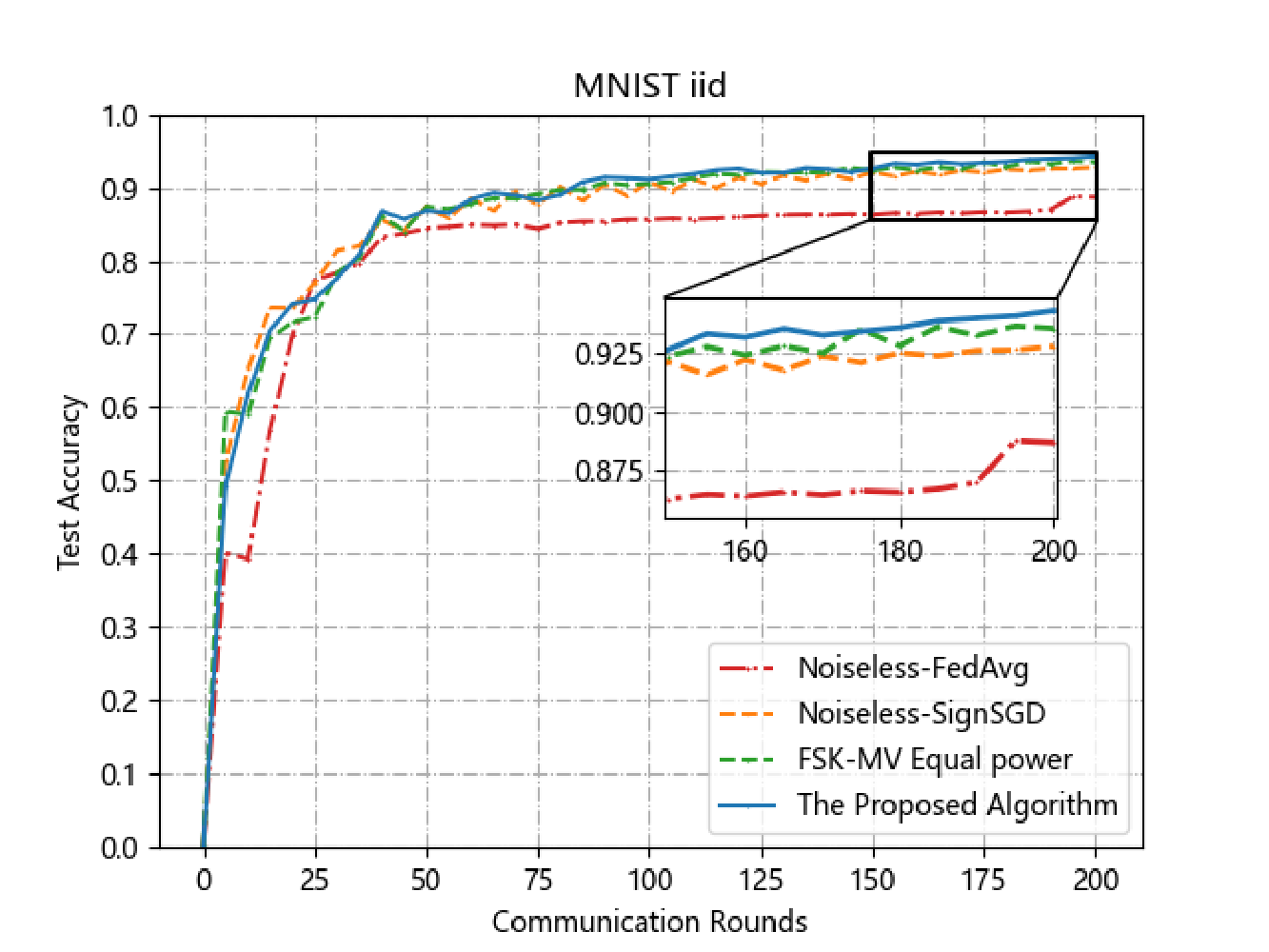}}
\caption{Test accuracy versus communication round under iid setting on the MNIST dataset.}
\label{icml-historical}
\end{center}
\vskip -0.2in
\vspace{-1.0em}
\end{figure}

\begin{figure}[ht]
\vspace{-2.0em}
\vskip 0.2in
\begin{center}
\centerline{\includegraphics[width=\columnwidth]{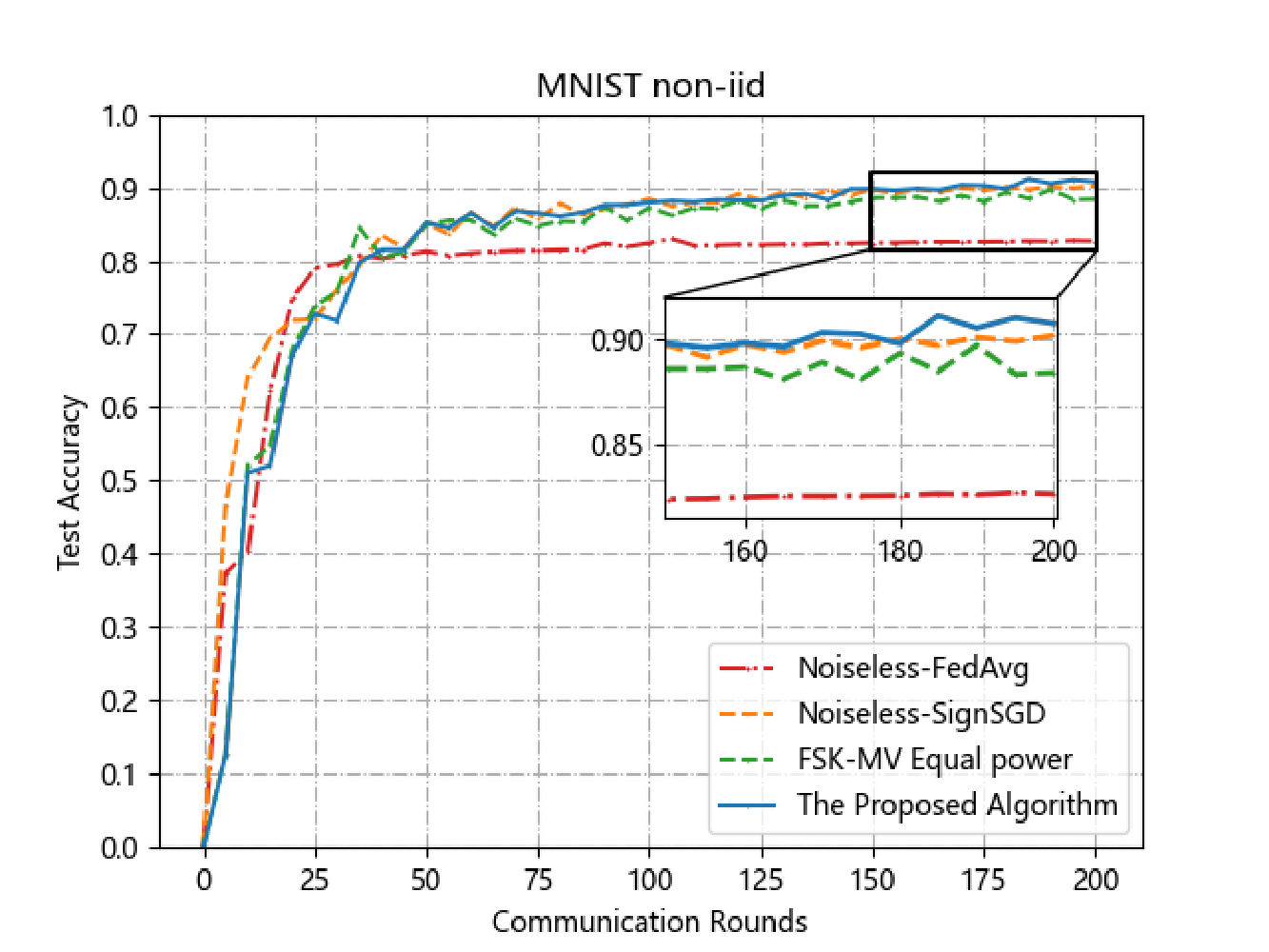}}
\caption{Test accuracy versus communication round under non-iid setting on the MNIST dataset.}
\label{icml-historical}
\end{center}
\vskip -0.2in
\vspace{-1.0em}
\end{figure}

In Figures 1 and 2, we provide the test accuracy results for iid/non-iid data by taking time-synchronization errors. For two baseline algorithms, simple noise-free aggregation is not otherwise optimized, and the aggregation of parameters leads to slightly worse error accuracy than the proposed algorithm. Also, the results about the proposed scheme and FSK-MV indicate that both have a high level of test accuracy with the time synchronization error. 

Due to the dynamic power control scheme, our test results are still superior to the FSK-MV even under the terrible environmental conditions. Also the usage of non-coherent detection causes high test accuracy without the utilization of CSI at the ED. The ultimate experimental results indicate that the whole scheme can mitigate the impact of time synchronization error, channel fading and noise.

\nocite{langley00}

\bibliography{example_paper}
\bibliographystyle{icml2023}

\newpage
\appendix
\onecolumn
\section{Training Process of  Federated Learning via Non-Coherent Over-the-air Computation}
\begin{figure}[ht]
\vskip 0.2in
\begin{center}
\centerline{\includegraphics[width=\columnwidth]{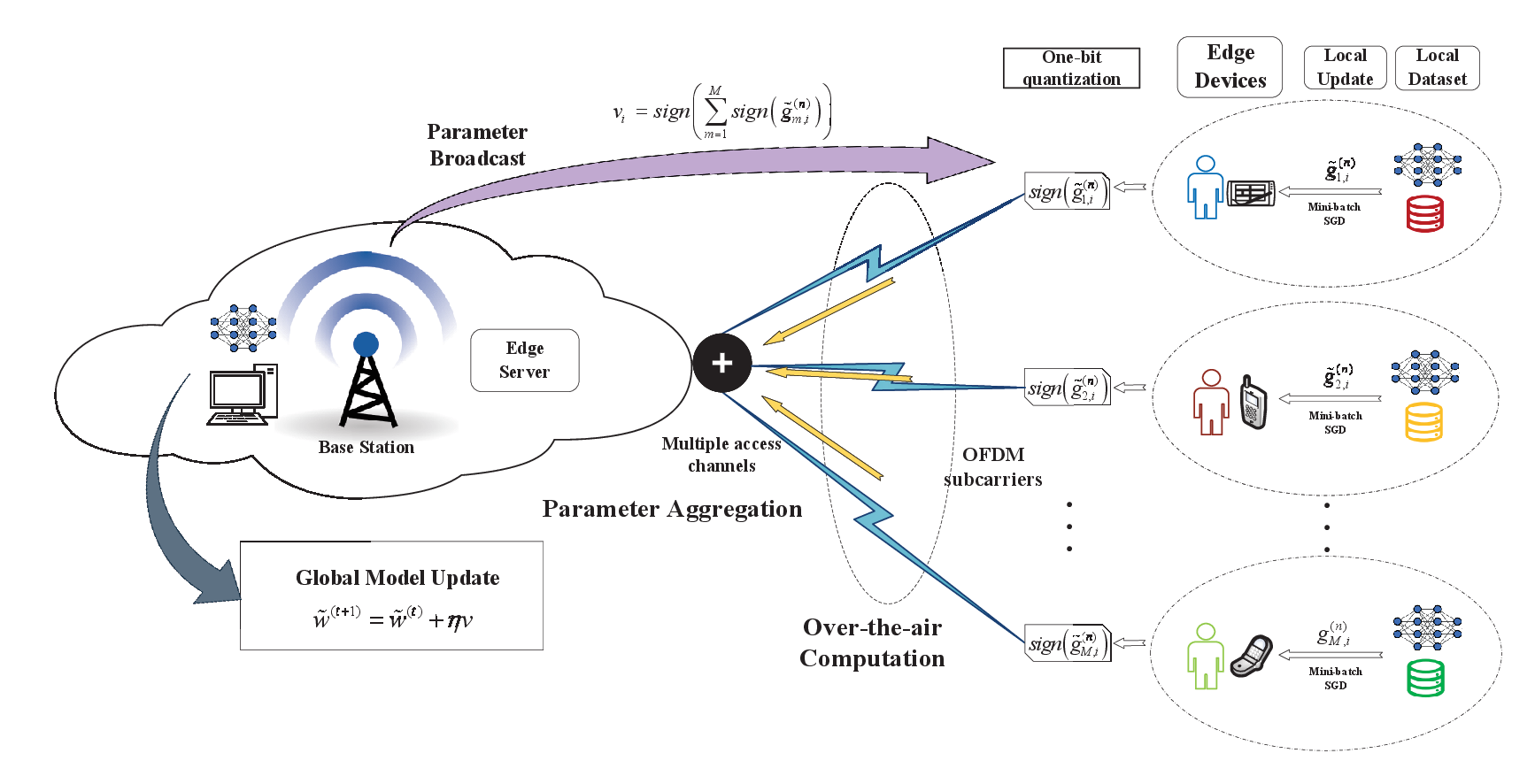}}
\caption{The FL system of the proposed scheme comprising of $M$ EDs coordinated by an ES.}
\label{icml-historical}
\end{center}
\vskip -0.2in
\end{figure}

\section{Details of Non-Coherent Energy Detection}

After exploiting the modulation and power control scheme at
the transmitter side, we complete the symbol transmission through the Aircomp scheme. Then, we can obtain the model parameters for uplink aggregation at the receiver ES side and we assume that CSI is not available at this time.

It is known that $\Delta_{i}^{(n)} = \mathbf{e}_{i}^{+} - \mathbf{e}_{i}^{-}$ is the sum of the energy values for which the detection vote is 1 or -1, where $e_{i}^{+} \triangleq|r_{l^{+},m^{+}}^{(n)}|_{2}^{2}$ and $e_{i}^{-} \triangleq|r_{l^{+}, m^{+}}^{(n)}|_{2}^{2}$ , $\forall i$. At the receiver side, it is not expected to recover the sign of the local stochastic gradient, since we need the vote summation, and do not exploit any method to resolve the interference introduced by the channel. Instead, the choice is to exploit the interference for aggregation and to compare the energy magnitude on two different subcarriers to detect the MV in (11). 

Here the interference usage for aggregation means that we incorporate the transmitting power, noise power and other channel factors together into the subcarriers energy magnitude. This operation is equivalent to using the interference term as the standard content of the detection, and then performing an one-bit quantization scheme.

\textbf{The Proof of Lemma 3.1:} According to the above description, the specific proof details are obtained as follows. We assume that the multipath channels between the  EDs and the ES are independent. To simplify the notation, we omit the index $n$. Since (9) is a weighted summation of independent complex Gaussian random variables with zero mean and unit variance (i.e., channel coefficients), $t_{l^{+}, m^{+}}^{(n)}$ is a zero mean random variable, where its variance is
\begin{equation} 
\begin{split}
\mu_{i}^{+}=\mathbb{E}\left[e_{i}^{+}\right] & =\mathbb{E}\left[\left|r_{l^{+}, m^{+}}^{(n)}\right|_{2}^{2}\right]=\mathbb{E}\left[E_{0}\sum_{\bar{s}_{\forall m, i}^{(n)}=1}{{P}_{m}}+\sigma_{\mathrm{n}}^{2}\right] \\
& =E_{0} M_{i}^{+} \mathbb{E}\left[{{P}_{m}}\right]+\sigma_{\mathrm{n}}^{2}=E_{0} M_{i}^{+} \vartheta+\sigma_{\mathrm{n}}^{2}.
\end{split}
\end{equation}
The same analysis can be done for $\mu_{i}^{-}$.

As given in Lemma 3.1, $\mu_{i}^{+}$ and $\mu_{i}^{-}$ are linear functions of $\emph{M}_{i}^{+}$ and $\emph{M}_{i}^{-}$. Also, we obtain the correct MV because the symbols energy may not coherently add up. However, the detection performance depends on the parameter $\vartheta$ that has an efficient trade-off about the direction of convergence of the global model on $\emph{e}_{i}^{+}$ and $\emph{e}_{i}^{-}$.

\section{Assumptions for The Subsequent Analysis}
In this work, we consider the well-known Lipschitz continuity utilized to explain some of the assumptions. Moreover, in order to make the developed theory applicable to neural networks rather than assuming a convex loss function, we require a lower bound. Specially, it is worth noting that the minimum assumption required to guarantee convergence to the stabilization point.

\begin{definition}
A function $\emph{f}$ is $\emph{L}$-Lipschitz over a set $\emph{s}$ with respect to a norm $\|\cdot\|$ if there exist a real constant $\emph{L}>0$ such
that $\|f(\mathbf{y})-f(\mathbf{x})\| \leq L\|\mathbf{y}-\mathbf{x}\|, \forall \mathbf{x}, \mathbf{y} \in S$.
\end{definition}

\begin{lemma}
(Lemma 1.2.3 \cite{allenzhu2018natasha}). For a differentiable function $f:\mathbb{R}^{Q}\rightarrow\mathbb{R}$, let $\nabla f$ be $\emph{L}$-Lipschitz on $\mathbb{R}^{Q}$ with respect to norm $\|\cdot\|_{2}$. Then, for any $\mathbf{y}, \mathbf{y}$ from $\mathbb{R}^{Q}$,
\begin{equation}
\left|f(\mathbf{y})-f(\mathbf{x})-\nabla f(\mathbf{x})^{\mathrm{T}}(\mathbf{y}-\mathbf{x})\right| \leq \frac{L}{2}\|\mathbf{y}-\mathbf{x}\|_{2}^{2}.
\end{equation}
\end{lemma}

\begin{assumption}
(Bounded Loss Function). For all parameter vectors $\mathbf{w}$, the lower bound of the associated loss function is some value $F^{*}$, $F(\mathbf{w}) \geq F^{*}, \forall \mathbf{w}$.
\end{assumption}

Assumptions C.3 and C.4 as follow, on the Lipschitz smoothness and bounded variance, respectively, are standard in the stochastic optimization literature.

\begin{assumption}
(Smoothness). Let $\mathbf{g}$ denote the gradient of the loss function $F(\mathbf{w})$ evaluated at $\mathbf{w}$. For all $\mathbf{w}$ and $\mathbf{w'}$, the expression from (21) is given by
\begin{equation}
|F(\mathbf{w}^{\prime})-(F(\mathbf{w})-\mathbf{g}^{\mathrm{T}}(\mathbf{w}^{\prime}-\mathbf{w}))| \leq \frac{1}{2} \sum_{i=1}^{q} L_{i}\left(w_{i}^{\prime}-w_{i}\right)^{2},
\end{equation}
\end{assumption}
where we can assume that there exists a vector of non-negative constants $\mathbf{L}=\left[L_{1}, \ldots, L_{q}\right]^{\mathrm{T}}$.

\begin{assumption}
(Variance bound). The stochastic gradient estimates ${\tilde{\mathbf{g}}_{k}=\left[\tilde{g}_{k, 1}, \ldots, \tilde{g}_{k, q}\right]^{\mathrm{T}}=\nabla F_{k}\left(\mathbf{w}^{(n)}\right)}, \forall k$ are independent and unbiased estimates of  $\mathbf{g}\!=\![\mathbf{g}_{1}, \ldots,\mathbf{g}_{q}]^{\mathrm{T}}\!=\!\nabla F(\mathbf{w})$ (the true gradient) with a coordinate bounded variance, i.e.,
\begin{equation}
\mathbb{E}\left[\tilde{\mathbf{g}}_{k}\right]=\mathbf{g}, \forall k,
\end{equation}
\begin{equation}
 \mathbb{E}\left[\left(\tilde{g}_{k, i}-g_{i}\right)^{2}\right] \leq \sigma_{i}^{2} / d_{\mathrm{b}}, \forall k, i,
\end{equation}
\end{assumption}

where $\mathbf{\sigma}=\left[\sigma_{1}, \ldots, \sigma_{q}\right]^{\mathrm{T}}$ is a non-negative constant vector,  $\tilde{g}_{k, i}$ and $g_{i}$ denote the $\emph{i}$th element of $\tilde{\mathbf{g}}_{k}$ and $\mathbf{g}$.

After these analysis above, another significant assumption is that the data-stochasticity induced gradient noise. Meanwhile, this assumption causes the discrepancy between $\tilde{\mathbf{g}}_{k}$ and $\mathbf{g}$, which is unimodal and symmetric as verified by experiments in \cite{bernstein2018signsgd} and formally described as follow.

\begin{assumption}
(Unimodal, symmetric gradient noise). For any
given $\mathbf{w}$, each elements of the vector
$\tilde{\mathbf{g}}_{k}, \forall k,$ has a unimodal
distribution that is also symmetric around its mean.
\end{assumption}

At this time, it can be obviously appreciated that Gaussian noise is a special case. Noting that even for small batches of other magnitudes, we expect the central limit theorem to be in effect and to bring the typical gradient noise distribution close to a Gaussian distribution.

\begin{assumption}
(Independent, identical, and unbiased gradients). The local stochastic gradient estimates are independent
and unbiased, i.e., $\mathbb{E}_{\tilde{\mathcal{D}}_{m}}\left[\tilde{g}_{m, i}^{(t)}\right]=g_{i}^{(n)}, \forall m, i$.
\end{assumption}

\begin{assumption}
(Exponential probability distribution). For given $\emph{M}_{i}^{+}$ and $\emph{M}_{i}^{-}$, $e_{i}^{+}$ and $e_{i}^{-}$ are exponential random variables, where their means are $\mu_{i}^{+}$ and $\mu_{i}^{-}$, respectively.
\end{assumption}

Assumption C.7 does not claim that the local stochastic gradients are unbiased estimates of the global gradients. Therefore, they are accommodated to heterogeneous data distribution scenarios where the sum of local stochastic gradients is unbiased.

\section{The proof of Lemma 3.2}

We consider the establishment of equivalent mathematical events $\operatorname{sign}(\Delta_{i}^{(n)})=\operatorname{sign}(g_{i}^{(n)})$ described by a well-defined random variable with known distribution. To address this, assume that $\operatorname{sign}(g_{i}^{(n)})=1$ as a standard. Let $\emph{X}$ denote the number of edge devices with correct sign at the $\emph{i}$th element of the gradient vector, namely, with $\operatorname{sign}(\tilde{g}_{m, i}^{(n)})=\operatorname{sign}(g_{i}^{(n)})$, i.e., $\operatorname{sign}(g_{i}^{(n)})=1$. For the scenario, the random variable $\emph{X}$ can then be model as the sum of $\emph{K}$ independent Bernoulli trials, and binomial with success probability and failure probability denoted by:
\begin{equation}
p_{i} \triangleq \mathbb{P}\left[\operatorname{sign}\left(\tilde{g}_{m, i}^{(n)}\right)=\operatorname{sign}\left(g_{i}^{(n)}\right)\right],
\end{equation}
and
\begin{equation}
q_{i} \triangleq \mathbb{P}\left[\operatorname{sign}\left(\tilde{g}_{m, i}^{(n)}\right) \neq \operatorname{sign}\left(g_{i}^{(n)}\right)\right],
\end{equation}
where $p_{i}+q_{i}=1$. They are intuitively determined by the randomness of the data.

\textbf{The Proof of Error Probability:} Regarding the bounds on stochasticity-induced error, we mainly deal with the term of $P_{i}^{\mathrm{err}}$. Through the previous descriptions, the following treatment is available for (14), for all $m$. This implies that
\begin{equation}
P_{i}^{\mathrm{err}}=\sum_{M_{i}^{+}=0}^{M} \mathbb{P}\left[\operatorname{sign}\left(\Delta_{i}^{(n)}\right) \neq 1 \mid X=M_{i}^{+}\right] \mathbb{P}\left[X=M_{i}^{+}\right].
\end{equation}

According to the properties of Bernoulli distribution, the second term on the right side of the equation can be expressed as follow: 
\begin{equation}
\mathbb{P}\left[X=K_{i}^{+}\right]=\left(\begin{array}{c}
K \\K_{i}^{+}\end{array}\right) p_{i}^{K_{i}^{+}} q_{i}^{K-K_{i}^{+}}.
\end{equation}

To calculate $\mathbb{P}\left[\operatorname{sign}(\Delta_{i}^{(n)}) \neq 1 \mid X=M_{i}^{+}\right]$, we need to derive the probability problem in dynamic power control, which can be shown as
\begin{equation}
\begin{array}{l}
P\left(\operatorname{sign}(\Delta_{i}) \neq \operatorname{sign}(g_{i}^{(n)})\right)=p_{i}^{\text {err }},\\
P\left(\operatorname{sign}(\Delta_{i})=\operatorname{sign}(g_{i}^{(n)})\right)=1-p_{i}^{\text {err }},\\
P\left(\operatorname{sign}(\tilde{g}_{k, i}^{(n)})\neq \operatorname{sign}(q_{i}^{(n)})\right)=q_{i}, \\
P\left(\operatorname{sign}(\tilde{g}_{k, i}^{(n)})=\operatorname{sign}(g_{i}^{(n)})\right)=1-q_{i}.
\end{array}
\end{equation}

Also, we can obtain the probability of the dynamic power control term as $\left[(1-p_{i}^{e r})\left(1-q_{i}\right)\right]-\left(p_{i}^{err} \cdot q_{i}\right)=(1-p_{i}^{err}-q_{i})$. Based on the similarity analysis in \cite{9771881}, it is simple to prove that $e_{i}^{+}$ and $e_{i}^{-}$ are exponential random variables. And we derive the dynamic power control term in terms of energy detection (i.e., probabilistic representation) as
\begin{equation}
\mathbb{P}\left[\operatorname{sign}(\Delta_{i}^{(n)}) \neq 1 \mid X=M_{i}^{+}\right]=\frac{\left(1-P_{i}^{err}-q_{i}\right) K_{i}^{-} E_{0}\vartheta+\sigma_{n}^{2}}{K E_{0}\vartheta+2 \sigma_{n}^{2}}.
\end{equation}

Then, further combining the terms in (27) to simplify, an upper bound on $P_{i}^{\mathrm{err}}$ can be completed as
\begin{equation}
\begin{aligned}
P_{i}^{err} \leq &\sum_{K_{i}^{+}=0}^{K}\frac{K_{i}^{-}(1-q_{i}) E_{0}\vartheta+\sigma_{n}^{2}}{K E_{0} \vartheta+2 \sigma_{n}^{2}} \cdot \left(\begin{array}{l}K \\ K_{i}^{+}\end{array}\right) p_{i}^{K_{i}^{+}} q_{i}^{K-K_{i}^{+}}
\\=&\sum_{K_{i}^{+}=0}^{K}\frac{\left(K-K_{i}^{+}\right) \cdot p_{i} E_{0}\vartheta+\sigma_{n}^{2}}{K E_{0} \vartheta+2 \sigma_{n}^{2}} \cdot\left(\begin{array}{l}
K \\
K_{i}^{+}
\end{array}\right) p_{i}^{K_{i}^{+}} q_{i}^{K-K_{i}^{+}}.
\end{aligned}
\end{equation}

Analysis of (28) shows that the right-hand side of the inequality can be split into two parts and solved for separately by utilizing the properties of binomial coefficients.

(1) The-first-part can be obtained as
\begin{equation}
\begin{aligned}
\sum_{K_{i}^{+}=0}^{K}&\frac{\left(K-K_{i}^{+}\right) p_{i} E_{0} \vartheta}{K E_{0} \vartheta+2 \sigma_{n}^{2}}\left(\begin{array}{c}
K \\
K_{i}^{+}
\end{array}\right) p_{i}^{K_{i}^{+}} q_{i}^{K-K_{i}^{+}} \\
= & \frac{K \cdot p_{i}}{K+2 / \beta}\left(p_{i}+q_{i}\right)^{K}-\sum_{K_{i}^{+}=0}^{K} \frac{p_{i} \cdot K_{i}^{+}}{k+2 / \beta}\left(\begin{array}{c}
K \\
K_{i}^{+}
\end{array}\right) p_{i}^{K_{i}^{+}} q_{i}^{K-K_{i}^{+}} \\
= & \frac{K p_{i}}{K+2 / \beta}-\frac{p_{i} E\left(K_{i}^{+}\right) }{K+2 / \beta}
= \frac{K p_{i}\left(1-p_{i}\right)}{K+2 / \beta}
= \frac{K \left(1-q_{i}\right) q_{i}}{K+2 / \beta}.
\end{aligned}
\end{equation}

(2) The-second-part can be obtained as
\begin{equation}
\begin{aligned}
\sum_{K_{i}^{+}}^{K} & \frac{\sigma_{n}^{2}}{K E_{0} \vartheta+2 \sigma_{n}^{2}}\left(\begin{array}{c}
K \\
K_{i}^{+}
\end{array}\right) p_{i}^{K_{i}^{+}} q_{i}^{K-K_{i}^{+}} \\
& =\frac{1 / \beta}{K+2 / \beta}\left(\begin{array}{c}
K \\
K_{i}^{+}
\end{array}\right) p_{i}^{K_{i}^{+}}q_{i}^{k-k_{i}^{+}} \\
& =\frac{1 / \beta}{K+2 / \beta}\left(p_{i}+q_{i}\right)^{K}=\frac{1 / \beta}{K+2 / \beta}.
\end{aligned}
\end{equation}

Thus, this upper bound on $P_{i}^{\mathrm{err}}$ can be obtained after combining (29) with (30) as
\begin{equation}
P_{i}^{err} \leq \frac{K q_{i}\left(1-q_{i}\right)+1 / \beta}{K+2 \beta}.
\end{equation}

Expecting to obtain a more accurate upper bound, we further scale for $q_{i}\left(1-q_{i}\right)$, and an established fact is $q_{i}<1/2$.
\begin{equation}
\begin{array}{l}
y(a)=a(1-a)-b a \quad\left(0<a<\frac{1}{2}, 0<b<1\right),
\end{array}
\end{equation}

where $a$ is substituted with $q_{i}$, and b represents the slope of the function. Then we need to process (32) with $y^{\prime}(a)=(1-b)-2 a=0$, and get $a=(1-b)/2 \in\left(0, \frac{1}{2}\right)$. Under the analysis of function monotonicity, it is necessary to ensure that $y(a)>0$ holds, which gives $0<b<1/2$. Ultimately, we can rewrite (31) in the form as follow:
\begin{equation}
P_{i}^{err} \leq \frac{K q_{i}/2}{K+2 / \beta}+\frac{1 / \beta}{K+2 / \beta}.
\end{equation}

To proceed with, we need a bound on $q_{i}$ that can be
associated with the signal-to-noise ratio of a component of the stochastic gradient as $S_{i}$, which is defined in Lemma 3.2. 

Under the unimodal symmetric gradient noise assumption mentioned in Assumption C.6, we can obtain the following bound on $q_{i}$. The failure probability for the sign bit of a single ED is exploited by the following Lemma D.1.
\begin{lemma}
(Failure probability under conditions of unimodal symmetric gradient noise). Based on several previous assumptions, the failure probability satisfies:
\begin{equation}
\begin{aligned}
q_{i} & =\mathbb{P}\left[\operatorname{sign}\left(\tilde{g}_{k, i}^{(n)}\right) \neq \operatorname{sign}\left(g_{i}^{(n)}\right)\right]\\
& \leq\left\{\begin{array}{ll}
\frac{2}{9} \frac{1}{R_{i}^{2}}, & \text { if } R_{i}>\frac{2}{\sqrt{3}} \\
\frac{1}{2}-\frac{R_{i}}{2 \sqrt{3}}, & \text { otherwise, }
\end{array} \right.
\end{aligned}
\end{equation} 
which is less than $1/2$ for all cases. 
\end{lemma}
The proof process relies on the properties of certain probability distributions, which is captured in Appendix E. Under the symmetry assumption, by exploiting the derivations in Lemma D.1, $q_{i} \leq \frac{\sqrt{2} \sigma_{i}}{3|g_{i}^{(n)}| \sqrt{n_{\mathrm{b}}}}$ still holds true. Then, we combine the upper bound on $q_{i}$ with (33) to complete the proof of Lemma 3.2.

\section{The Proof of Lemma D.1}
\textbf{The Proof of Failure Probability:} Under the background of Assumption C.5 and Assumption C.6, for a unimodal symmetric random variable $Y$ with mean $\mu$ and variance $\sigma^{2}$, the following Gauss’ inequality holds:
\begin{equation}
\begin{aligned}
\mathbb{P}[|Y-\mu|>y] \leq\left\{\begin{array}{ll}
\frac{4}{9} \frac{\sigma^{2}}{y^{2}}, & \text { if } \frac{y}{\sigma}>\frac{2}{\sqrt{3}} \\
1-\frac{y}{\sqrt{3} \sigma}, & \text { otherwise. }
\end{array}\right.
\end{aligned}
\end{equation}

Then applying symmetry followed by Gauss’ inequality, the failure probability can be obtained by
\begin{equation}
\begin{aligned}
\mathbb{P} & \left[\operatorname{sign}(\tilde{g}_{k, i}^{(n)}) \neq  \operatorname{sign}(g_{i})\right]  =\mathbb{P}\left[\tilde{g}_{k, i}^{(n)}-g_{i} \geq|g_{i}|\right] \\
& =\frac{1}{2} \mathbb{P}\left[|\tilde{g}_{k, i}^{(n)}-g_{i}| \geq|g_{i}|\right] \\
& \leq\left\{\begin{array}{ll}
\frac{2}{9} \frac{\sigma_{i}^{2}}{d_{b}|g_{i}^{(n)}|^{2}}, & \text { if } \frac{|g_{i}^{(n)}|}{\sigma_{i} / \sqrt{n_{b}}}>\frac{2}{\sqrt{3}} \\
\frac{1}{2}-\frac{|g_{i}^{(n)}|}{2 \sqrt{3} \sigma_{i} / \sqrt{d_{b}}}, & \text { otherwise. }
\end{array}\right.
\end{aligned}
\end{equation}

Eventually, we complete the proof, which is utilised to infer Lemma 3.2.

\section{The Proof of Theorem 3.4}
Note that the OFDM symbols may non-coherently add up and their amplitudes may not be aligned in fading channel. Hence, the MV calculated in (11) is different from the original MV given in (3). Then, we provide a complete overview of the convergence performance. Firstly, we define the convergence rate \cite{9272666} as the rate at which the expected value of average norm of the gradient of $F(\mathbf{w})$ diminishes as the number of total communication rounds $\emph{N}$ and $\emph{M}$, when the training is done in the presence of the proposed scheme.

\textbf{The Proof of Convergence Rate Analysis:} The proof is carried out following widely-adopted strategy of the norm of gradient with respect to the expected improvement made in a single algorithm step. And we compare this with total possible improvement under Assumption C.3. 

To begin with, we firstly bound the improvement of the objective for the data-stochasticity induced noise based on Assumption C.4. For processing, we exploit the contents of (4) in our inference process as a way to decompose (the data and the channel) stochasticity-induced error that we need to analyze. Thus, by utilising Assumption C.4 and (4), we can write as
\begin{equation}
\begin{aligned}
F(\mathbf{w}^{(n+1)}) & -F(\mathbf{w}^{(n)}) \leq \mathbf{g}^{(n)^{\mathrm{T}}}\left(\mathbf{w}^{(n+1)}-\mathbf{w}^{(n)}\right)\\&
+\frac{1}{2} \sum_{i=1}^{q} L_{i}\left(w_{i}^{(n+1)}-w_{i}^{(n)}\right)^{2}. 
\end{aligned}
\end{equation}
Then, we exploit (4) to make a substitution with $(\tilde{v}_{i}^{(n)})^2=1$ whether it is $+1$ or $-1$. Therefore, we have
\begin{equation}
\begin{aligned}
F(\mathbf{w}^{(n+1)}) & -F(\mathbf{w}^{(n)}) \leq-\eta \mathbf{g}^{(n)^{\mathrm{T}}}\mathbf{\tilde{v}}^{(n)}+\frac{1}{2} \sum_{i=1}^{q} L_{i}(\eta\tilde{v}_{i}^{(n)})^2 \\
= & -\eta\left\|\mathbf{g}^{(n)}\right\|_{1}\operatorname{sign}\left(\Delta_{i}^{(n)}\right)+\frac{\eta^{2}}{2}\|\mathbf{L}\|_{1}. \\&
\end{aligned}
\end{equation}
The first term on the right side of the equation can be analyzed to know that the value of $\operatorname{sign(\cdot)}$ cannot be determined, so the term will have a randomness error. Thus, we then proceed to obtain as
\begin{equation}
\begin{aligned}
F(\mathbf{w}^{(n+1)}) & -F(\mathbf{w}^{(n)}) \leq-\eta \mathbf{g}^{(n)^{\mathrm{T}}} \mathbf{v}^{(n)}+\frac{\eta^{2}}{2}\|\mathbf{L}\|_{1} \\
= & -\eta\|\mathbf{g}^{(n)}\|_{1}+\frac{\eta^{2}}{2}\|\mathbf{L}\|_{1} \\
& +2 \eta \sum_{i=1}^{q}|\mathbf{g}_{i}^{(n)}| \mathbb{I}\left[\operatorname{sign}\left(\Delta_{i}^{(n)}\right) \neq \operatorname{sign}\left(g_{i}^{(n)}\right)\right].
\end{aligned}
\end{equation}

Thus, the expected improvement of the left term can be written as an inequality as follows.
\begin{equation}
\begin{aligned}
\mathbb{E}\left[F(\mathbf{w}^{(n+1)})\right. & \left.-F(\mathbf{w}^{(n)}) \mid \mathbf{w}^{(n)}\right] \leq-\eta\|\mathbf{g}^{(n)}\|_{1}+\frac{\eta^{2}}{2}\|\mathbf{L}\|_{1} \\
+ & \underbrace{2 \eta \sum_{i=1}^{q}|\mathbf{g}_{i}^{(n)}| \underbrace{\mathbb{P}\left[\operatorname{sign}\left(\Delta_{i}^{(n)}\right) \neq \operatorname{sign}\left(g_{i}^{(n)}\right)\right]}_{\triangleq P_{i}^{err}}}_{\text {Stochasticity-induced error }}.
\end{aligned}
\end{equation}

For the analysis of the above equation, the main challenge is to obtain an upper bound on the error term in (40). The bound is a function of the stochasticity of the local gradients and the detection performance of the proposed scheme.

Accordingly, based on Lemma 3.1 and several definitionins Theorem 3.4, an upper bound on the stochasticity-induced error can be represented by the proof related to Appendix D as follows:
\begin{equation}
\begin{aligned}
\sum_{i=1}^{q}|g_{i}^{(n)}| p_{i}^{err} \leq  & \sum_{i=1}^{q}|g_{i}^{(n)}|\frac{\frac{K}{2} \cdot \sqrt{2} / (3 R_{i})}{K+2 / \beta}\\+&\sum_{i=1}^{q} | g_{i}^{(n)}|\frac{1 / \beta}{K+2 / \beta}.
\end{aligned}
\end{equation}

Then, substituting Lemma 3.2 into (41), this specific upper bound is written as 
\begin{equation}
\begin{aligned}
\sum_{i=1}^{q}|g_{i}^{(n)}| p_{i}^{err} \leq
&\sum_{i=1}^{q} \frac{\frac{K}{2}\cdot|g_{i}^{(n)}|}{K+2 / \beta} \cdot \frac{\sqrt{2}|\sigma_{i}^{(n)}|}{3|g_{i}^{(n)}| \sqrt{d_{b}}} \\+&\sum_{i=1}^{q}|g_{i}^{(n)}| \frac{1 / \beta}{K+2 / \beta}.
\end{aligned}
\end{equation}

Now, after a series of collations, we can obtain as follow:
\begin{equation}
\begin{aligned}
\sum_{i=1}^{q}|g_{i}^{(n)}| p_{i}^{err} \leq
\frac{K}{K+2 / \beta} \cdot \frac{\sqrt{2}}{6 \sqrt{d_{b}}}\|\sigma\|_{1}+\frac{1 / \beta}{K+2 / \beta}\|g^{(n)}\|_{1}.
\end{aligned}
\end{equation}

Then, under considering Assumption C.3, we perform a telescoping sum over the iterations and calculate the expectation over the randomness in the trajectory as
\begin{equation}
\begin{aligned}
F&(\mathbf{w}^{(0)})-F^{*} \geq F(\mathbf{w}^{(0)})-F(\mathbf{w}^{(N)})
\\=&\mathbb{E}\left[\sum_{n=0}^{N-1} F(\mathbf{w}^{(n)})-F(\mathbf{w}^{(n+1)})\right]\\
\geq & E\left[\sum_{n=0}^{N-1}\left(\left(\eta-2 \eta \cdot \frac{1 / \beta}{K+2 / \beta}\right)\|g^{(n)}\|_{1}-\frac{\eta^{2}}{2}\|L\|_{1} \right.\right. \\&\left.\left. -\frac{2 \eta \cdot K}{K+2 / \beta} \cdot \frac{\sqrt{2}}{6 \sqrt{d_{b}}} \cdot \| \sigma\|_{1}\right)\right]
\\=& E\left[\sum_{n=0}^{N-1}\left(\frac{K \eta \cdot \|g^{(n)}\|_{1}}{K+2 / \beta} -\frac{\eta^{2}\|L\|_{1}}{2}-\frac{2\sqrt{2} K \eta \cdot \|\sigma\|_{1}}{6 (K+2 / \beta)\sqrt{d_{b}}}\right)\right].
\end{aligned}
\end{equation}

In order to derive the term of required convergence rate, we rearrange (44) and use the expressions for $d_{b}$ and $\eta$, while conducting a series of simplifications to obtain as follow:
\begin{equation}
\begin{aligned}
E \left[\frac{1}{N} \sum_{n=0}^{N-1}\|g^{(n)}\|_{1}\right]
& \leq(1+\frac{2}{K \beta}) \frac{\sqrt{\gamma}}{2 \sqrt{N}}\cdot \sqrt{\| L \|_{1}}\\+&
(1+\frac{2}{K \beta}) \cdot \frac{\sqrt{\|L\|_{1}}\sqrt{N}}{N \sqrt{\gamma}}(F(w^{(0)})
-F^{*}(w))+\frac{2 \sqrt{2} \sqrt{\gamma} \|\sigma\|_{1}}{6 \sqrt{N}}).
\end{aligned}
\end{equation}

Finally, by replacing pertinent term with some expressions in Theorem 3.4, the proof process is completed and (16) is reached.

\section{The Relevant Conclusions of Convergence Rate}

Based on the above derivation, proof process and the visual representation of Theorem 3.4, we can infer the followings:
\begin{itemize}
\item In connection with a larger SNR (i.e., a larger $1/\sigma_{n}^{2}$ ) and a large number of EDs (i.e., a larger $\emph{K}$), the convergence rate with FSK-MV in fading channel improves since $\tau$ decreases.
\item The dynamic power control account for a better convergence in this scenario rate since $\vartheta$ increases with a state. It is higher equivalence degree of the sign between edge device gradient parameters and aggregation gradient parameters.
\item Under ideal power control, the convergence rate becomes similar to SignSGD in an ideal channel [\cite{8870236}, Theorem 1] gradually.
\end{itemize}
Note that the proposed scheme has no impact in terms of
convergence, since the interference of channel fading and noise is used for energy aggregation by the usage of a non-coherent detection scheme at the receiver side.

\section{Performance Comparison about The Proposed Scheme}
The discussion in the above part of this section has adequately addressed the actual performance performance regarding such problems. And there is a stark contrast related to the problems that can be solved in the uplink and downlink communications of FL. The communication issues solved are mainly focused on the following two areas:
\begin{itemize}
\item \textbf{Resistance to Time-Varying Fading Channel:} In contrast to the approaches in \cite{8870236}, the proposed scheme does not utilize the CSI for TCI at the EDs. It is therefore compatible with time-varying channels and does not lose the gradient information that we need during transmission owing to TCI. Because of the advantages mentioned above, a shown trade-off is that it quadruples the amount of time-frequency resources required in AirComp compared to OBDA in \cite{9272666}. More importantly, with addition of dynamic power control, interference from fading channels and noise can be addressed effectively.

\item  \textbf{Resistance to Time-Synchronization Errors:} It is worth noting that our holistic comprehensive scheme greatly enhances the resistance capability to time-synchronization errors. The obvious reason is that the time misalignment among the EDs or the uncertainty in receiver synchronization within the CP window lead to phase rotation in the frequency domain. while the comprehensive scheme does not encode information on the amplitude or phase, it does not use any channel-dependent information in the EDs and ES. Therefore, the considered process is more robust to time-synchronization errors compared to OBDA scheme.
\end{itemize}

\end{document}